\documentclass[sn-mathphys-num]{sn-jnl}% Math and Physical Sciences Numbered Reference Style 
\usepackage{graphicx}%
\usepackage{multirow}%
\usepackage{amsmath,amssymb,amsfonts}%
\usepackage{amsthm}%
\usepackage{mathrsfs}%
\usepackage[title]{appendix}%
\usepackage{xcolor}%
\usepackage{textcomp}%
\usepackage{manyfoot}%
\usepackage{booktabs}%
\usepackage{algorithm}%
\usepackage{algorithmicx}%
\usepackage{algpseudocode}%
\usepackage{listings}%
\usepackage{amssymb}
\usepackage{diagbox}
\usepackage{multicol}
\usepackage{multirow}
\usepackage{booktabs}
\usepackage{array}
\usepackage{makecell}
\usepackage{multirow}

\usepackage{rotating} 
\usepackage{bbding}

\theoremstyle{thmstyleone}%
\theoremstyle{thmstyletwo}%

\theoremstyle{thmstylethree}%

\begin{document}

\title[Article Title]{Mask-Guided Multi-Task Network for Face Attribute Recognition}

\author[1]{Gong Gao}
\author[1]{Zekai Wang}
\author[1]{Jian Zhao}
\author[2]{Ziqi Xie}
\author[1]{Xianhui Liu*}
\author[1]{Weidong Zhao}

\affil[1]{\orgdiv{School of Computer Science and Technology}, \orgname{Tongji University},
                \city{Shanghai},
                \postcode{200092},
                \state{Shanghai},
                \country{China}}
\affil[2]{ \orgname{Shanghai University of Engineering Science},
                \city{Shanghai},
                \postcode{201620},
                \state{Shanghai},
                \country{China}}
% \cortext[1]{Xianhui Liu}


\abstract{Face Attribute Recognition (FAR) plays a crucial role in applications such as person re-identification, face retrieval, and face editing. Conventional multi-task attribute recognition methods often process the entire feature map for feature extraction and attribute classification, which can produce redundant features due to reliance on global regions. To address these challenges, we propose a novel approach emphasizing the selection of specific feature regions for efficient feature learning. We introduce the Mask-Guided Multi-Task Network (MGMTN), which integrates Adaptive Mask Learning (AML) and Group-Global Feature Fusion (G2FF) to address the aforementioned limitations. Leveraging a pre-trained keypoint annotation model and a fully convolutional network, AML accurately localizes critical facial parts (e.g., eye and mouth groups) and generates group masks that delineate meaningful feature regions, thereby mitigating negative transfer from global region usage. Furthermore, G2FF combines group and global features to enhance FAR learning, enabling more precise attribute identification. Extensive experiments on two challenging facial attribute recognition datasets demonstrate the effectiveness of MGMTN in improving FAR performance.}




\maketitle

\section{Introduction}
With the advent of deep learning, FAR has become a fundamental component in applications such as face retrieval \cite{suo2024knowledge,huang2024attribute}, person re-identification \cite{wu2024attributes,ahmed2025multi}, and attribute editing \cite{ghani2024securing,han2024face,ding2025stable}, owing to its efficiency and practicality. Facial attributes, typically encompassing a diverse set of labels, are naturally amenable to feature extraction via Multi-Task Networks (MTNs) \cite{hekimoglu2024active,wu2025dynamic}.
The MTN framework facilitates simultaneous learning of multiple related tasks by sharing convolutional feature extractors among tasks, while employing group-specific modules with non-shared parameters to generate group-specific features. However, reliance on global features may result in feature redundancy, and the use of inappropriate non-shared group-specific modules can lead to negative transfer, ultimately impairing performance \cite{li2024cspformer}.
\begin{figure}[!htpb]
\centering
\includegraphics[width=1.0\linewidth]{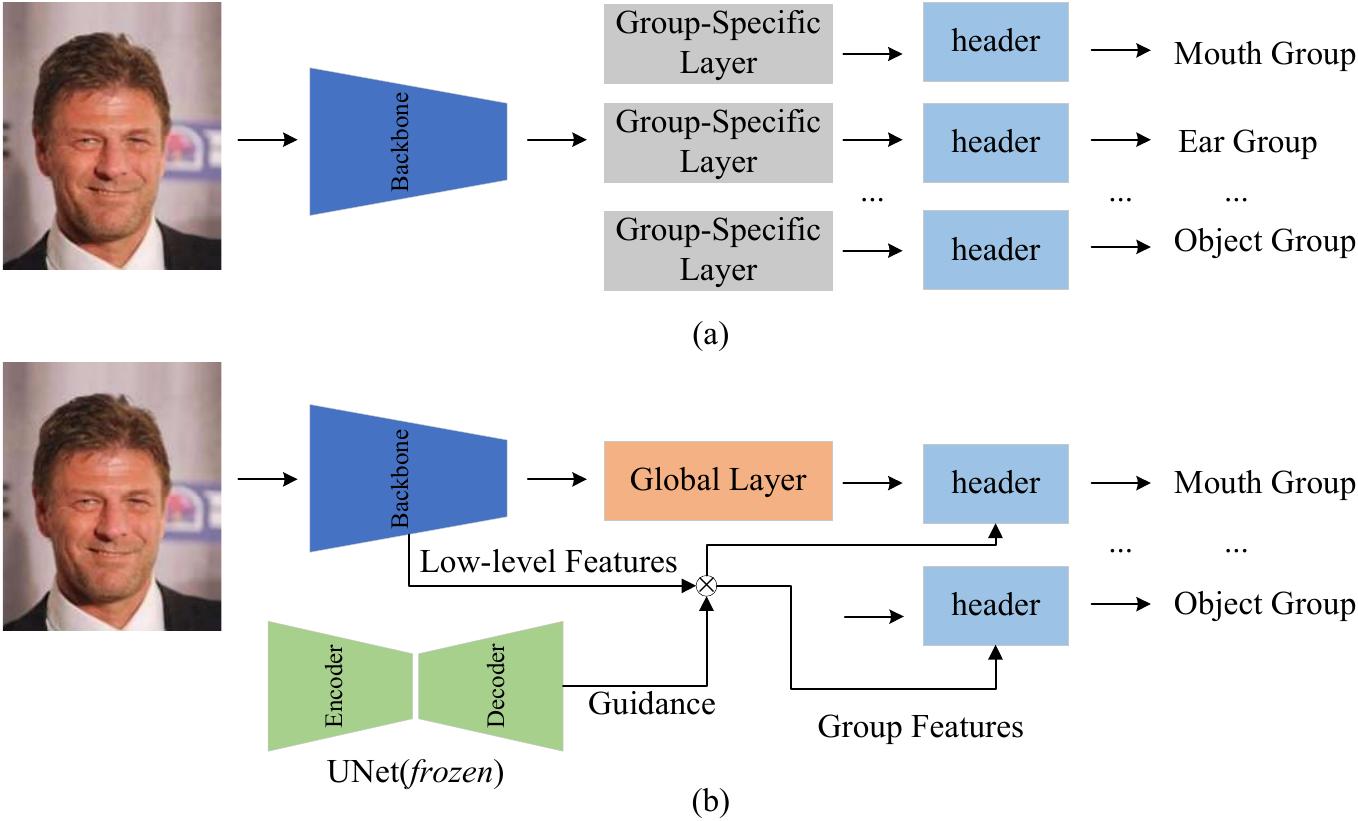}
\caption{\label{fig:0}  
(a) The conventional global-based approach prior to multi-task attribute recognition. 
(b) The MGMTN approach, which leverages masks to guide group feature learning, effectively reduces feature redundancy typical of global-based methods. Global layers extract holistic features, while UNet generates group features from low-level feature maps. Here, $\otimes$ denotes element-wise multiplication.
}

\end{figure}

Early studies \cite{zhang2014panda,kalayeh2017improving} focused on manually segmenting facial regions and extracting features using techniques such as Histogram of Oriented Gradients (HOG) \cite{dalal2005histograms}, Scale-Invariant Feature Transform (SIFT) \cite{lowe2004distinctive}, or Local Binary Pattern (LBP) \cite{ojala1994performance}, followed by classification through Support Vector Machine (SVM) \cite{cortes1995support}. However, these methods heavily depend on precise segmentation and often exhibit limited effectiveness with rotated faces in extreme poses. To minimize manual intervention and leverage the advanced feature extraction capabilities of CNNs, researchers have begun to directly identify attributes from images.
The advent of powerful CNNs negates the need for manual engineering of feature vectors, unlike hand-crafted features \cite{zhang2014panda}. However, many deep feature regions are either irrelevant or nonsensical, highlighting the necessity to identify useful regions. Recent advancements have shown the potential of deep region selection in enhancing multi-task learning \cite{wei2025modeling} and benefiting multi-task attribute recognition.

\begin{figure}[!htpb]
\centering
\includegraphics[width=1.0\linewidth]{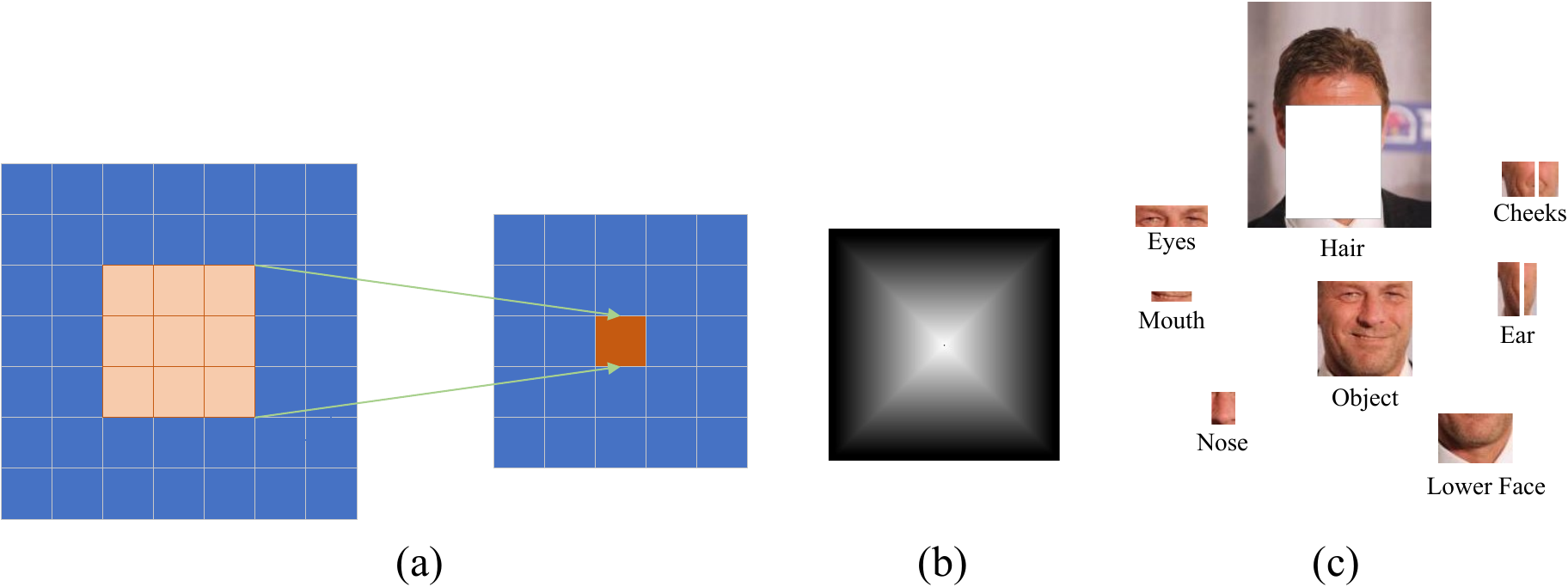}
\caption{\label{fig:7}  
(a) A $3\times3$ convolution generates a feature map by aggregating information from 9 neighboring pixels into a single feature value.  
(b) In ResNet50 or ResNeSt50, features extracted by a convolutional layer exhibit decreasing similarity with increasing spatial distance from each location.  
(c) Comparison across various facial parts and object locations, retaining only the regions of interest.  
}

\end{figure}
Some researchers have proposed group-based methods to learn specific group features with multiple non-shared parameters, as shown in Figure \ref{fig:0} (a).
Chen \emph{et al.} \cite{chen2021improving}, Mao \emph{et al.} \cite{mao2020deep}, and Gao \emph{et al.} \cite{gao2025far} introduced MTN that employs parameter-sharing strategies to adaptively explore semantic relationships between attributes. 
Researchers have also concentrated on enhancing feature representation through the use of multi-scale feature information. Studies by Zheng \emph{et al.} \cite{zheng2020blan}, Song \emph{et al.} \cite{song2022prior}, and Chen \emph{et al.} \cite{chen2023learning} have leveraged backbone networks to capture multi-level features, enabling the utilization of global context. These methods integrate low-level texture, color, and shape features with high-level semantic features to enhance feature representation. Moreover, Chen \emph{et al.} \cite{chen2021improving} have approached attribute inference by utilizing features from various scales of features, instead of directly classifying attributes from features extracted solely by the last layer of CNNs. Therefore, a limitation of their methods is that the group-specific modules learn features for the entire facial region, which leads to feature redundancy and the risk of negative transfer \cite{zhou2023feature}.

In Convolutional Neural Networks (CNNs), we observe that as the spatial distance between pixels in an image increases, the feature information gradually diminishes, as illustrated in Figure \ref{fig:7} (a)(b).
Moreover, due to the utilization of a large amount of task-irrelevant redundant features, negative transfer is caused. 
In FAR, labels correspond to distinct regions for feature extraction, and utilizing global feature regions for prediction often leads to negative transfer. The intuitive notion is to crop out distracting regions, which proves beneficial for identifying specific facial attributes within corresponding regions, as illustrated in Figure \ref{fig:7} (c). Numerous studies \cite{zhang2014panda,ehrlich2016facial,kalayeh2017improving,ding2018deep,zheng2022general,10064142} have validated this approach, while pixel-level masking methods \cite{kalayeh2017improving,zheng2022general} rely more heavily on accurate predictions, with little emphasis on overlapping regions between groups.
% The approach to FAR described previously emphasizes global feature regions, leading to feature redundancy. 
% As illustrated in Figure \ref{fig:0}(b), a more effective strategy involves selecting deep feature regions that represent component information.
% This method contrasts with the reliance on global feature regions (i.e., the entire feature map) as shown in Figure \ref{fig:0}(a), facilitating more targeted learning and reducing redundancy. 

To address the challenges presented, we introduce MGMTN that incorporates AML and G2FF. This approach aims to effectively manage both the extraction of correspondence features from specific regions and their subsequent enhancement. By categorizing face attributes through intuitive analysis and creating group masks with a UNet \cite{hu2024perspective}, and employing ResNeSt \cite{zhang2022resnest} to derive multi-scale features, we isolate specific group deep regions through AML. Then, G2FF realizes the complementary strengths of both group and global features. Attribute prediction for each group is performed using fully connected layers.

Our key contributions can be summarized in threefold:
\begin{itemize}
\item MGMTN is designed to mitigate negative transfer in the context of multi-task attribute recognition.
\item AML enables the identification of discriminative regions for each facial attribute group. By directing group-specific modules to focus on relevant regions, AML reduces redundant features and minimizes negative transfer. Additionally, G2FF is leveraged to effectively integrate the group feature representations obtained via AML.
\item The proposed framework offers valuable insights into multi-task attribute recognition, demonstrating its advantages. Empirical results on public datasets validate the competitiveness of our approach.
\end{itemize}
\section{Relate Works}
In this section, we provide a comprehensive review of multi-task attribute recognition, highlighting its significance and recent advancements. Subsequently, we offer a concise overview of contemporary research on AG methods and mask generation techniques. This discussion establishes the context for our proposed approach, situating it within the broader landscape of related work in FAR.
\subsection{Multi-Task Attribute Recognition}

In recent years, the integration of multi-task learning into attribute recognition has yielded substantial performance gains. Within the domain of FAR, a variety of feature extraction strategies have been explored, including global-based methods \cite{hand2017attributes,rudd2016moon,han2017heterogeneous}, local-based approaches \cite{zhang2014panda,shu2021learning}, network architecture–driven techniques \cite{fang2017dynamic,lu2017fully,li2021auto, Chennas}, and feature fusion–oriented methods \cite{chen2021improving,chen2023learning,song2022prior}, among others.
% 人脸属性识别中基于全局的方法采用多任务框架进行特征提取，随后采用属性分组方法进行属性识别。

Global-based approaches initially employ an AG method, followed by a multi-task framework for feature extraction and attribute recognition. For instance, Deep Multi-Task Learning (DMTL) \cite{han2017heterogeneous} uses AlexNet as the backbone for estimating heterogeneous attributes. Multi-Channel Neural Network-Auxiliary (MCNN-AUX) \cite{hand2017attributes} integrates MTN with an auxiliary network to examine correlations. MOON \cite{rudd2016moon} and MTCNN \cite{he2019mtcnn}devise a hybrid objective optimization function for attribute learning, incorporating weighted loss for each attribute. Identity-Conditioned Knowledge Distillation (ICKD) \cite{10064142} accounts for both identity information and attribute relationships, utilizing identity information to boost face attribute estimation performance. These methodologies employ global features and utilize separate group-specific modules for each attribute group. Nevertheless, such approaches often result in feature redundancy, with overlapping information redundantly learned across various attribute classifiers, potentially diminishing the overall efficiency of FAR.

The local-based approach focuses on extracting specific facial feature regions pertinent to group attributes, coupled with subsequent feature extraction for attribute classification. This domain has seen the development of several prominent methods. For instance, PANDA \cite{zhang2014panda} employs pose alignment to ensure the consistency of facial images, which are then selectively cropped to highlight regions associated with attributes, facilitating attribute classification. SSPL \cite{shu2021learning} introduces three auxiliary tasks: Patch Rotation Task (PRT), Patch Segmentation Task (PST), and Patch Classification Task (PCT). DP-FAR \cite{he2018harnessing} leverages a Generative Adversarial Network (GAN) to create abstract images. These images, alongside the original facial images, are used to construct a dual-path face attribute recognition network. These methodologies utilize the spatial region information of facial images within a self-supervised learning framework. However, their effectiveness is often limited by the reliance on manual cropping or fine-grained segmentation supervision from pre-trained models, affecting their effectiveness.

Certain researchers have concentrated on the development of fine-grained networks to more effectively capture attribute features. This involves crafting networks with dynamic characteristics, such as dynamic width \cite{liu2024differentiable}, dynamic depth \cite{alturki2022depth}, or Neural Architecture Search (NAS) \cite{wang2024advances} for network design.
Fang \emph{et al.} \cite{fang2017dynamic} implemented a dynamic depth approach to facilitate early stopping, thus minimizing redundancy. Lu \emph{et al.} \cite{lu2017fully} introduced a network with dynamic width that forms connections among attributes, thereby enabling the modeling of their relationships for enhanced semantic linkage exploitation. While models with dynamic width and depth have shown performance enhancements, they necessitate meticulous configuration. NAS methods, as applied by Li \emph{et al.} \cite{li2021auto}, Chennas \emph{et al.} \cite{Chennas}, automate the architecture discovery process but require significant validation time, posing a risk of overfitting. Nevertheless, these techniques demand extensive computational resources due to the need for thorough exploration of the structural space.
% Inevitably, this necessitates a substantial amount of validation time, and these methods often result in overfitting.
% Inevitably, the above methods require a significant amount of time to search the space structure, which makes them computationally expensive.

Other researchers have explored the use of feature fusion techniques to accentuate fine-grained details, thereby enhancing overall recognition performance. Approaches such as APS \cite{chen2023learning} and MGG-Net \cite{chen2021improving} combine shared foundational features with group-specific sub-networks for adaptive feature extraction from these shared sub-networks. An efficient attention mechanism, through a multi-feature soft alignment module, significantly improves network performance. Similarly, MFM \cite{song2022prior} leverages adaptive attention for merging features from adjacent layers and hierarchically integrating features across different scales. These strategies effectively enhance group features, contributing to substantial improvements in recognition performance.
However, it is crucial to note that fusing global multi-scale features can lead to feature redundancy, which may cause the negative transfer of certain attributes. This redundancy can impair the efficiency of attribute recognition systems, underscoring the need for strategies that address and mitigate this issue.

\subsection{Attribute Grouping Methods}
The efficacy of AG in enhancing multi-task FAR methods has been well documented. Researchers have increasingly utilized grouping methods to streamline MTN models. MCNN-AUX \cite{hand2017attributes} grouped facial attributes based on semantic categories, including gender, Nose, Mouth, Eyes, Face, Head, Facial Hair, Cheeks, and Fat. In contrast, DMM-CNN \cite{mao2020deep} categorized attributes into objective and subjective groups. PS-MCNN-LC \cite{cao2018partially} divided attributes according to their positional relationships into Upper, Middle, Lower, and Whole Image Groups. 
% Further advancements were made with the APS  \cite{chen2023learning} and MGG-Net \cite{chen2021improving}, which refined the grouping method by focusing on partial and object-specific regions of interest within facial attribute features.

In this study, drawing inspiration from MGG-Net \cite{chen2021improving} and APS \cite{chen2023learning},  we refine the AG method by investigating both partial and object-specific regions associated with distinct facial attributes. As detailed in Table \ref{tab:1}, we categorize the regions of interest into seven part-based groups (Mouth, Ear, Lower Face, Cheeks, Nose, Eyes, and Hair), supplemented by an additional group dedicated to object-related attributes (Object), culminating in a comprehensive division into eight distinct groups.
 \begin{table}[htbp]
   \centering
     \begin{tabular}{p{6.95em}<{\centering}p{21.55em}<{\centering}p{5.5em}<{\centering}}
      \hline
      Group name &{Attributes} &Number of attributes  \\
     \hline
     Mouth & 5 o Clock Shadow, Big Lips, Mouth Slightly Open, Mustache, Wearing Lipstick, No Beard &6 \\
     Ear & Wearing Earrings&1\\
     Lower Face & Double Chin, Goatee, Wearing Necklace, Wearing Necktie&4 \\
     Cheeks & High Cheekbones, Rosy Cheeks, Sideburns &3 \\
     Nose  & Big Nose, Pointy Nose&2 \\
     Eyes  & Arched Eyebrows, Bags Under Eyes, Bushy Eyebrows, Narrow Eyes, Eyeglass&5 \\
     Hair  & Bald, Bangs, Black Hair, Blond Hair, Brown Hair, Gray Hair, Receding Hairline, Straight Hair, Wavy Hair, Wearing Hat&10 \\
     Object & Attractive, Blurry, Chubby, Heavy Makeup, Male, Oval Face, Pale Skin, Smiling, Young&9 \\
     \hline
     \end{tabular}%
     \caption{Group of 40 face attributes.}
   \label{tab:1}%
 \end{table}%

\subsection{Mask Generation}

\begin{figure}[htbp]
\centering
\includegraphics[width=1.0\linewidth]{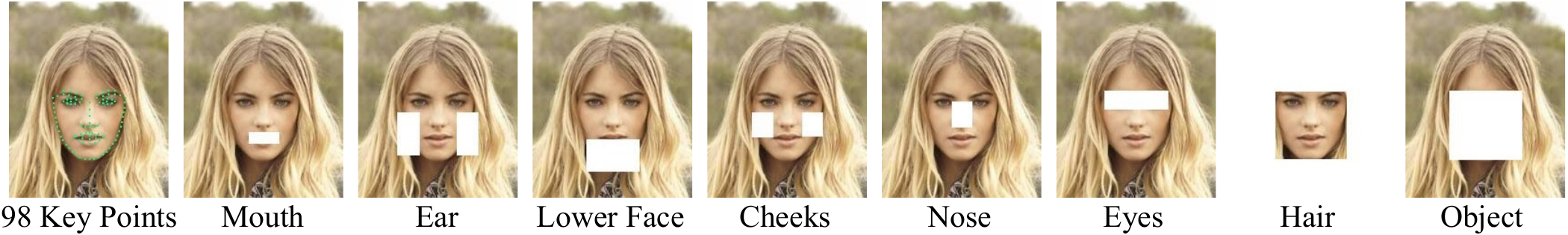}
\caption{\label{fig:1}  
Using keypoint annotations, rectangular contours are generated to delineate individual facial groups. Each mask highlights the corresponding group region with white pixels, while all other pixels represent the background.  
}

\end{figure}

As highlighted in previous discussions, the selection of informative deep features often receives insufficient attention in the realm of multi-task learning. Yet, identifying relevant regions while eliminating noisy or interfering elements remains essential, especially in multi-task learning contexts. Drawing on the spatial relationships depicted in Figure \ref{fig:7}(a)(b), we employ a masking method to mitigate interference from features associated with different attribute groups. In this context, ``region" pertains to a spatial mask (see Figure \ref{fig:1}) that highlights activated features within the convolutional layer.

% Recent progress in deep region selection is exemplified by Mask-CNN \cite{wei2018mask}, which addresses the intricacies of extracting fine-grained features \cite{ren2023focus}. Within this framework, specific deep regions are deemed valuable, while others are classified as background or noise. Leveraging this region selection strategy, Mask-CNN demonstrates enhanced efficacy in tasks related to fine-grained attribute recognition. In our study, we advance the mask generation approach employed by Mask-CNN by incorporating UNet, a more sophisticated mask generation model. UNet not only improves the efficiency of localizing the entire facial region but also aids in the precise supervision of the localization for each component group. This advancement results in a notable increase in the accuracy of component localization. Importantly, our MGMTN showcases the pivotal role of prior spatial knowledge in facilitating multi-task attribute recognition, underscoring its critical importance in this field.

\section{Methods}
\begin{figure*}[!htpb]
\centering
\includegraphics[width=1.0\linewidth]{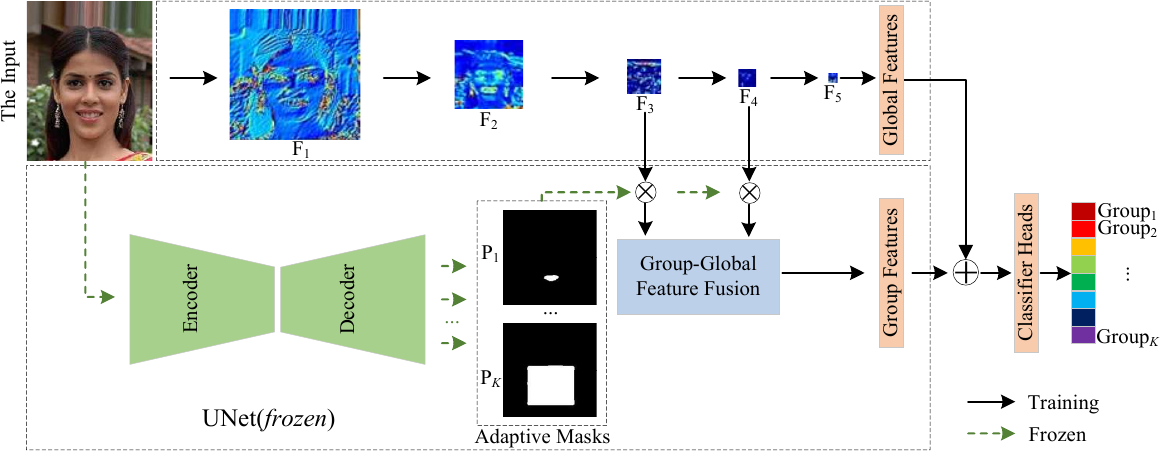}
\caption{\label{fig:3}The MGMTN architecture incorporates multiple groups, with some processing the entire feature map and others extracting specific group features. AML is employed to identify informative regions via group masks, which are then used to learn group features. These features are fused through the G2FF mechanism to produce the final group attribute representation. Each classification head consists of two fully connected layers: the first reduces the dimension of the enhanced group features $\rm{G}$ from 3584 to 512, and the second independently predicts attribute values for each group feature. Here, $\otimes$ denotes element-wise multiplication, and $\oplus$ denotes channel-wise concatenation.}

\end{figure*}
% These masks serve two purposes: locating the position of faces in the image and, more importantly, selecting the regions of interest for facial features. Leveraging UNet enables precise localization and efficient region selection.

In this section, we introduce the MGMTN, which consists of two primary components. The first component leverages a UNet, a fully convolutional network architecture, to generate partial masks. The second component implements a group feature fusion module, which utilizes the generated masks to obtain group-level features.

This scheme not only enables joint training but also captures effective spatial information at both the object and partial levels. By integrating attributes from multiple groups, MGMTN is capable of processing various attributes concurrently, thus improving the overall attribute recognition performance. The foundation of the MGMTN model is established through the combination of masks generated by UNet and the fusion of global and group features. This method emphasizes region selection and thorough attribute capturing, facilitating efficient multi-task learning in FAR.

% We first introduce the MGMTN model, comprising UNet for mask generation and a recognition network leveraging the masks. This joint training scheme captures effective spatial information at both object and partial levels. By combining multiple group attributes, MGMTN can concurrently handle diverse tasks, improving overall facial attribute recognition.

 \subsection{Adaptive Masks Learning}

To generate mask annotations, we utilize the pre-trained FaRL \cite{zheng2022general} model to annotate facial keypoints. These annotations are subsequently transformed into mask labels, which are utilized for training the UNet. This process imparts prior spatial knowledge through group masks that encapsulate each attribute group. 

% UNet is specifically designed for pixel labeling tasks and can process input images of varying resolutions while producing corresponding output images of the same size. In our method, we employ UNet not only for object and component localization within group regions, but also for generating segmentation predictions used as object and component masks in the subsequent feature fusion process.

For the CelebA \cite{liu2015deep} and LFWA \cite{huang2008labeled} datasets, UNet generates masks for each group based on 98 keypoints identified by FaRL. Specifically, seven groups correspond to partial masks that cover the smallest rectangles around specific facial regions. The remaining group represents the object mask, covering the rectangle that encompasses all facial regions. In Figure \ref{fig:1}, the masks are illustrated as white regions corresponding to different parts of the face.
\begin{figure}[htbp]
\centering
\includegraphics[width=1.0\linewidth]{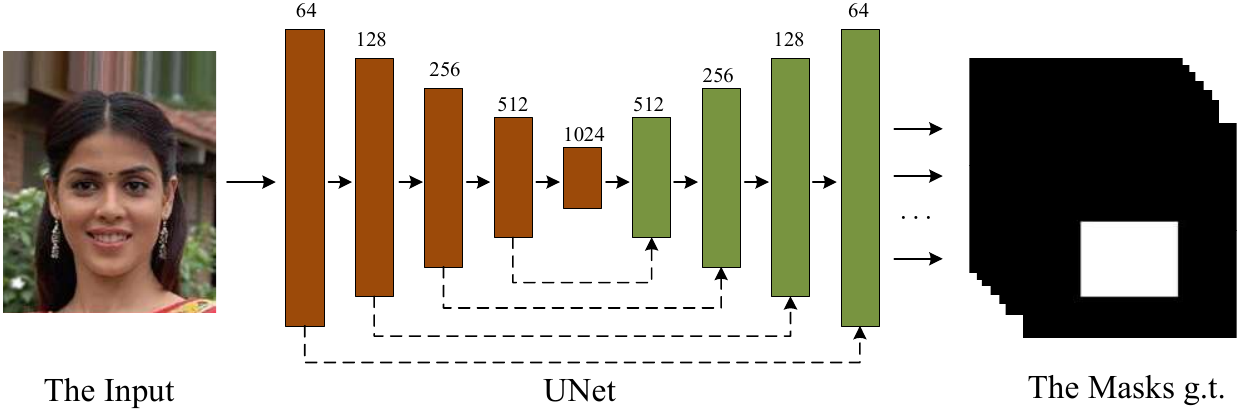}
\caption{\label{fig:2}Illustration of the AML process using UNet. The final convolutional layer of UNet consists of eight convolutional heads, each independently predicting a group mask.}

\end{figure}
% To ensure efficient training, all images in our dataset are maintained at their original resolution. 
To prepare inputs for the UNet, a scaling operation is applied to every raw image, resulting in an image block with dimensions of 227×227. The architecture of the AML network is depicted in Figure \ref{fig:2}.
Consistent with prior studies \cite{fang2023eva,wang2023masked}, we regard the bounding box-like group masks as ground truth for segmentation. Given the potential overlap among these regions, we approach the AML process as a multi-task learning problem. The prediction of group masks is treated as a multiple pixel-level binary classification task. UNet employs multiple prediction heads to generate predictions for group masks, each producing an output size of (227, 227, 1). 

The input to the global stream is the original image, resized to dimensions of 227×227×3. We employ ResNeSt50 to extract multi-scale features, denoted as ${\rm{F}} = \{ \rm{F_ \emph{i}}\}, \emph{i}=3,4,5$. Using the group masks ${\rm{P}} = \{ \rm{P_ \emph{j}}\}, \emph{j}=1,2,...,K$, obtained via AML—where $K$ denotes the total number of groups—we construct the MGMTN model for joint training. The learned group masks are then resized to 14×14 and 7×7 through downsampling.

For each group, we obtain a 14×14 region with a depth of 512 dimensions and a 7×7 region with a depth of 1024 dimensions from the deep convolution layers, respectively. 
% (as discussed in Section 3.1) 
These resized masks are subsequently employed to highlight the most relevant and discriminative feature regions. The selection process is performed by computing the element-wise product ($\otimes$) between the two corresponding regions, resulting in a multi-scale grouped feature representation ${\rm{Gr}}_i^j$. Formally, it can be defined as
\begin{equation}
{\rm{Gr}}_i^j = \text{resize}({\rm{P}}_j; {\rm{F}}_i) \otimes {\rm{F}}_i,
\end{equation}
where $\text{resize}({\rm{P}}_j; {\rm{F}}_i)$ denotes the operation that rescales the part mask ${\rm{P}}_j$ to match the spatial resolution of the feature map ${\rm{F}}_i$.

In our experiments, UNet is employed for group mask learning and prediction. During inference, no annotations are used, and each image produces 8 class heatmaps, maintaining the same size as the original input image.

% We randomly selected some qualitative examples of predicted part masks, as depicted in Figure 4. In these plots, the learned mask is overlaid on the original image. Masked parts are represented by white, while the background is black. It can be observed from these figures that although the ground-truth part masks exhibit some inaccuracy, the learned UNet yields more precise part masks. Additionally, these part masks efficiently locate the position of each part by enclosing it within closed rectangles. 
% Moreover, the grouped masks still demonstrate good segmentation of face parts, even without utilizing fine-grained processing techniques such as face segmentation networks (as shown in the third row of Figure 4). Quantitative results concerning part localization and object segmentation will be presented in Sections 4.3 and 4.4, respectively.

% However, there are certain failure cases, as demonstrated on the right side of Figure 4. In some instances, inaccuracy occurs when localizing regions with glasses or detecting regions near the ears. Furthermore, the torso mask may not be accurately predicted due to the size of the main object or the complexity of the background. For optimal recognition performance, precise predictions of the two partial masks would greatly benefit subsequent depth chart selection and final fine-grained classification. Therefore, we utilize the predicted masks for partial localization and feature region selection within MGMTN during both the training and testing phases.

 \subsection{Group-Global Feature Fusion}
Numerous studies \cite{kobayashi2019global,xie2023farp,cargan2024local} have shown that fusing global features with local features can significantly enhance feature complementarity.
Consequently, we employ a masked guidance approach to generate group features from lower-level features, which are then fused with higher-level, semantically rich features to create complementary features, thereby enhancing the expressiveness of the extracted features.

To refine the multi-scale group features ${\rm{Gr}}i$, we adopt a channel attention mechanism~\cite{hu2018squeeze} to enhance the discriminative representation of multi-scale features. This process begins by computing a global spatial descriptor through average pooling, which can be formulated as
\begin{equation}
Z_i = \frac{1}{{H \times W}}\sum\limits_{m = 1}^H {\sum\limits_{n = 1}^W {\rm{Gr_ \emph{i}}}}(m,n) 
\end{equation}
where $H$ and $W$ denote the height and width of the feature map, respectively.

For each group, we compute the channel attention vector ${\rm{\Psi}}_i \in \mathbb{R}^{1 \times C}$, where $C$ is the number of channels. The set of channel attentions is denoted as ${\rm{\Psi}} = {{\rm{\Psi}}_i}, i = 3, 4$. The attention weights are obtained as
\begin{equation}
{\rm{\Psi}}_i = \text{Sigmoid}\big(\text{Conv1}_i(Z_i)\big),
\end{equation}
where $\text{Conv1}_i$ represents a convolutional layer with stride $=1$ and padding $=0$, used to generate enhanced attention responses.

The refined feature maps are then computed by reweighting the original group features with the corresponding attention vector:
\begin{equation}
\tilde{\rm{Gr}}_i = {\rm{Gr}}_i \times {\rm{\Psi}}_i.
\end{equation}

For each multi-scale group, the refined features are concatenated to form a 1536-dimensional representation:
\begin{equation}
\tilde{\rm{Gr}} = \text{Concat}(\tilde{\rm{Gr}}_3, \tilde{\rm{Gr}}_4).
\end{equation}
Finally, the concatenated group features are fused with the global feature ${\rm{Gl}}$ to produce a 3584-dimensional feature vector:
\begin{equation}
{\rm{G}} = \text{Concat}(\tilde{\rm{Gr}}, {\rm{Gl}}) = \text{Concat}(\tilde{\rm{Gr}}_3, \tilde{\rm{Gr}}_4, {\rm{Gl}}).
\end{equation}
 \subsection{Model Training and Testing Phases}
The overall structure of our proposed model is illustrated in Figure \ref{fig:3}. To demonstrate the data flow, let us consider the full image stream as an example.
% Figure \ref{fig:3} illustrates the necessity of preserving the weight of the feature region when it falls within the target range, while it is crucial to discard its weight when the feature region is located in the background area. 
During our implementation, the mask comprises learned partial segmentation masks, depicted as an 8-channel matrix with binary values, where 1 signifies retention, and 0 signifies exclusion.

The selection mechanism operates through the element-wise multiplication of the convolutional activation tensor with the mask matrix. As a result, regions within the specified range are preserved based on their weights, whereas other regions are converted into 0-vectors. Should UNet classify pixels as belonging to regions of interest, the corresponding values in the mask are retained. In contrast, if pixels are identified as background, their corresponding values in the mask are set to 0, akin to the process used in salient object detection \cite{jia2023tfgnet}. 
% The processed mask then selects relevant regions for subsequent processing.

For the regions selected, max-pooling is applied to derive separate feature vectors. These vectors are then concatenated into 1536-dimensional (1536-d) features and subsequently integrated with global features. In our implementation, G2FF follows the AML process. The final layer of the MGMTN model comprises multiple classification layers dedicated to attribute recognition. The entire MGMTN model undergoes training, maintaining the parameters of the pre-trained UNet segmentation network fixed throughout the training phase. 

During inference, when a test image is introduced, the pre-trained UNet initially produces mask predictions for each group without requiring additional data.
%  \subsection{Model Testing Phases}

\section{Experimental Results and Analysis}

In this section, we describe the experimental setup and detail the implementation process. Following this, we present the recognition metrics obtained through our proposed method. Moreover, we assess the performance in terms of part and object segmentation. Lastly, we discuss the efficiency and effectiveness of the proposed MGMTN.

\subsection{Datasets and Metrics}
We assessed the MGMTN using two widely recognized benchmarks, CelebA and LFWA datasets, aligning with methodologies from other works in the domain of multi-task learning for attribute recognition.

{\bf CelebA.}
Originating from the Chinese University of Hong Kong, the CelebA dataset stands as a comprehensive collection of facial attributes within images, featuring cluttered backgrounds and notable variations in pose. It comprises 202,599 face images, each annotated with 40 distinct facial attributes, such as Bald, Black Hair, among others.

{\bf LFWA.}
The LFW dataset consists of 13,233 face images, each with a resolution of 250×250 pixels. This dataset is characterized by a wide range of pose variations, including extreme poses and occlusions. The LFWA dataset, an extension of LFW, includes 40 binary facial attribute labels, adopting the same attribute categorization introduced in the CelebA dataset.

{\bf Metrics.}
To evaluate the performance of the model, we adopt accuracy and average class-balanced mean accuracy (mA) \cite{wang2019dynamic} as metrics.
\subsection{Experimental Details}

We adhere to the standard train-test splits for both datasets. During the training phase, FaRL generates ground truth with partial masks using 98 keypoint annotations from each dataset. In the testing phase, there is no requirement for supervised signals, such as partial annotations or bounding boxes.

The implementation of the proposed MGMTN and UNet for mask generation utilizes the PyTorch \cite{paszke2019pytorch} open-source library. 
% After obtaining the learned group masks, we extract multi-scale group features.% as described in Section 3.2. 
The backbone network, ResNeSt50, is initialized through pre-training on ImageNet. To achieve convergence, we fine-tune the global features and each group individually before proceeding to joint training. For a fair comparison with other methods, we also implement the MGMTN model using ResNet50 as a basis. Data augmentation techniques, including horizontal flipping and random rotation, are applied to the training data. After fine-tuning each group, the MGMTN model undergoes joint training.
%, as depicted in Figure 2.

% During the testing phase, we extract the group features extraction region by utilizing UNet to obtain the correspondence mask for each group. We further enhance these features through the G2FF. Finally, we aggregate the global features and employ a fully-connected layer for attribute prediction, yielding the attribute prediction results for each group.

All experiments are conducted using the Adam optimizer \cite{kingma2014adam}, with an initial learning rate of 1e-4, momentum of 0.9, weight decay of 5e-4, and a batch size of 36. The classification head adopts Dropout with a rate of 0.5, and Equalized Focal Loss \cite{li2022equalized} is employed for each attribute.
For data preprocessing and augmentation, input images are first resized to $256\times256$ pixels and then randomly cropped to $227\times227$. To improve model generalization, random rotation (up to $\pm 15^\circ$) and horizontal flipping are applied with a probability of 0.35. These augmentations ensure both scale invariance and better generalization of the learned representation.
All experiments are carried out on a workstation equipped with an Intel Xeon E5-2678 v3 CPU, 64GB RAM, and two Nvidia GTX 2080 Ti GPUs.

% \begin{table}[htbp]
%    \centering
% \begin{tabular}{ccc}
% \hline
% \diagbox{Label}{Predict} & Positive & Negative \\
% \hline
% Positive & $TP$ & $FN$ \\
%     Negative &$FP$ & $TN$ \\
% \hline
% \end{tabular}
% \caption{Confusion Matrix.}
% \label{tab:2}%
% \end{table}

%  As can be seen in Table \ref{tab:2}, $TP$ and $TN$ are the number of correct predictions by the classifier, and $FP$ and $FN$ are the number of classification errors.

%  \begin{equation}
% \centering
%    \begin{aligned}
% {\rm{Accuracy}} &= \frac{{TP + TN}}{{Tp + TN + FP + FN}}\\
% {\rm{Precision}} &= \frac{{TP}}{{TP + FP}}\\
% {\mathop{\rm Recall}\nolimits} &= \frac{{TP}}{{TP + FN}}\\
% {\rm{F1}} &= \frac{{2*{\rm{Precision}}*{\rm{Recall}}}}{{{\rm{Precision}} + {\rm{Recall}}}} \\
% {\rm{m}}{{\rm{A}}_i} &= \frac{1}{2}(\frac{{T{P_i}}}{{T{P_i} + F{N_i}}} + \frac{{T{N_i}}}{{T{N_i} + F{P_i}}})\\
% {\rm{mA}} &= \frac{{\sum\limits_i^{|C|} {{\rm{m}}{{\rm{A}}_i}} }}{{|C|}}
% \end{aligned}
% \end{equation}
\subsection{Evaluation on the CelebA and LFWA Datasets}
\begin{table*}[htbp]
   \centering
  \resizebox{1.0\textwidth}{!}{
     \begin{tabular}{cccccc}
      
     \hline 
           \multirow{2}{*}{Methods}   &\multirow{2}{*}{Backbone} & \multicolumn{2}{c}{Accuracy/\%}
           &\multicolumn{2}{c}{mA/\%}\\ \cmidrule{3-6} 
      &&
      \multicolumn{1}{c}{CelebA} &\multicolumn{1}{c}{LFWA}&
     
      \multicolumn{1}{c}{CelebA} &\multicolumn{1}{c}{LFWA} \\
\hline
DMTL \cite{han2017heterogeneous}  &AlexNet&92.10 &86.00&-&- \\
% AttCNN \cite{hand2018doing}  &-&90.97 &73.03&-&-&-&- \\
DMM-CNN \cite{mao2020deep}   &ResNet50&91.70 &86.56&-&- \\
 % HFE \cite{yang2020hierarchical}  &ResNet50&92.17 & -&-&-&-&- \\
ICKD \cite{10064142}  &ResNet18&92.02 &86.77&-&- \\
MTCNN \cite{he2019mtcnn} &ResNet50&91.96 &\textbf{87.22}&88.82&84.56 \\

\hline
FFS-MTN \cite{lu2017fully}   &VGG16&91.26&83.24&-&- \\
% PS-MCNN-LC \cite{cao2018partially} &(2018CVPR)&-&92.22 & \textbf{87.36} \\
\hline
MGG-Net \cite{chen2021improving}  &ResNet50&92.00 &87.20&87.20&\textbf{85.19} \\
APS \cite{chen2023learning}&VGG16&92.12&86.74&-&-\\

\hline
 LNets+ANet \cite{liu2015deep}   &- &87.33&83.85&-&- \\
            10\%CelebA \cite{lingenfelter2021improving}  &ResNet18&91.81&-&82.39&-\\
             FaRL \cite{zheng2022general}  &ViT-B&91.88&86.69&-&-\\
            \hline
           
            SSPL \cite{shu2021learning}  &ResNet50&91.77 &86.53&-&- \\
DP-FAR \cite{he2018harnessing}  &ResNet50&91.81 &85.18&88.13&77.50 \\
               baseline &ResNeSt50&90.91 &85.62&88.01&80.72 \\
               MGMTN &ResNeSt50& \textbf{92.30} &87.01&\textbf{89.33}&81.10 \\
                MGMTN &ResNet50&92.19 &86.87&89.20&81.29 \\
           \hline
     \end{tabular}%
     }
          \caption{Comparison of experimental results between the MGMTN method and the latest approaches on the CelebA and LFWA datasets.}

   \label{tab:3}%
 \end{table*}%

We applied the MGMTN to perform experiments on the CelebA and LFWA datasets, utilizing the grouping method detailed in Table \ref{tab:1}.
 
As reported in Table \ref{tab:3}, MGMTN achieved state-of-the-art accuracy of 92.30\% and 89.33\% mA on the CelebA dataset. MGMTN leverages AML to identify regions for extracting group features, effectively reducing feature redundancy in multi-task attribute recognition. G2FF is used to realize the effective fusion of group features and global features.

We conducted comparisons with classic methods across various categories: global-based methods \cite{han2017heterogeneous,mao2020deep,10064142}, local-based methods \cite{he2018harnessing,shu2021learning}, network design-based techniques \cite{lu2017fully}, feature fusion-based methods \cite{chen2021improving} and other methods \cite{liu2015deep,lingenfelter2021improving}.
Among these, LNets+ANet, which utilizes a two-stage network for predicting face attributes, achieved accuracy of 87.33\% and 83.85\% on CelebA and LFWA, respectively. However, employing global face images for feature extraction and recognition in multi-attribute prediction tasks often results in negative transfer.

Both DMTL, DMM-CNN, and MTCNN have adopted AG and MTN for feature extraction. These methods rely on global features to directly predict facial attribute results, lacking prior knowledge about the regions of interest for complementary parts. 
% AttCNN introduced a selective learning method that improves accuracy by dynamically adjusting the weight of the loss for each attribute task. 
However, this strategy might lead to less optimal training results for certain attributes. In contrast to the intricate architectures of FFS-MTN, MGMTN effectively minimizes unnecessary background noise and achieves superior recognition performance by using masks to isolate group features.
HFE and ICKD incorporated identity information to boost attribute recognition performance. Nevertheless, this technique is not adaptable to many in-the-wild scenarios. 
MGG-Net implemented group and graph learning methods to capture multiple group features and facilitate independent predictions at different levels of the network. However, such independence in prediction may lead to overly coarse generalization, thereby limiting the effective exploitation of complementary multi-scale features.
Contrary to SSPL, FaRL, and DP-FAR, MGMTN does not necessitate extra fine-grained pixel-level supervision during the model inference phase or a significant amount of pretraining data. 
\subsection{Multi-Task Part Segmentation Performance}

%   \begin{figure}[t]
% \begin{center}
%    \includegraphics[width=1.0\linewidth]{4.pdf}
% \end{center}
% \caption{A random sample of predicted part masks from the test set on the CelebA dataset. The first column represents the input original image, while the subsequent columns illustrate the learned part masks. In these figures, the pixels predicted as background are visually denoted by the color black.}
% \label{fig:4}
% \end{figure}
To generate ground-truth annotations for facial part segmentation, we employ a pre-trained FaRL face alignment model on the CelebA and LFWA datasets.
The qualitative segmentation results are illustrated in Figure~\ref{fig:1}. Although our UNet-based segmentation module generally performs well in capturing foreground facial regions, it occasionally struggles to delineate fine-grained details such as eyebrows and boundary edges. Since facial part segmentation is not the primary focus of this work, we intentionally refrain from applying additional pre- or post-processing refinement techniques.

\begin{figure}[!htpb]
\centering
\includegraphics[width=1.0\linewidth]{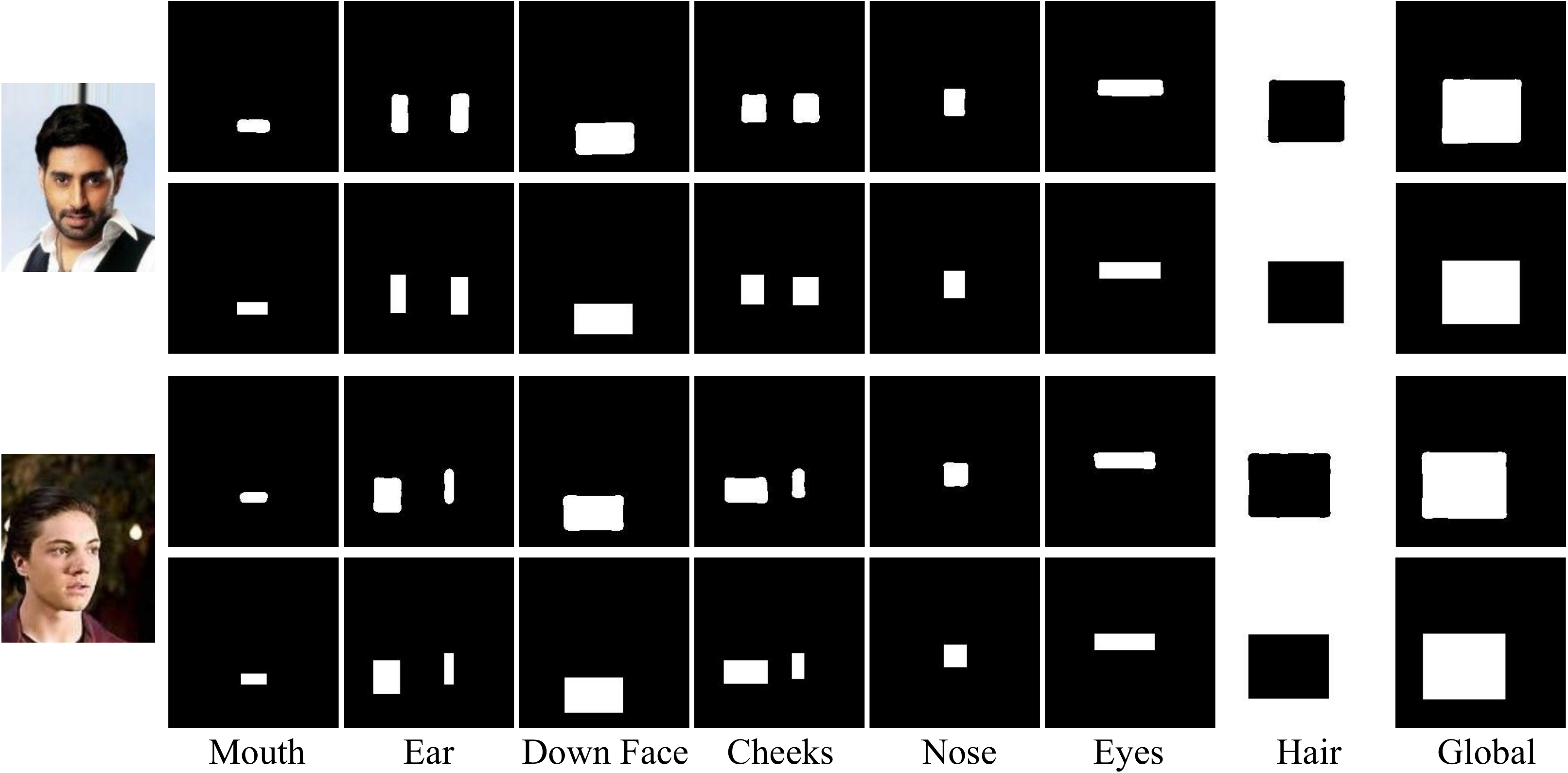}
\caption{\label{fig:6}A series of group mask visualizations generated from UNet inference, illustrating a comparison between model predictions and ground truth. The first and third rows present the predicted group masks, while the second and fourth rows show the corresponding ground truth.}
\end{figure}

The group masks predicted by UNet are shown in Figure~\ref{fig:6}.
For quantitative evaluation, we adopt the F1-measure, which assesses the overlap between the predicted mask and the ground-truth segmentation. On the test set, our method achieves an average F1 score of 89.02\%.
It is worth noting that improved segmentation quality directly enhances the accuracy of the derived partial masks, thereby contributing to better overall model performance.
As illustrated in Figure~\ref{fig:6}, the generated group masks consistently capture semantically meaningful facial regions, even under pose variations.

\subsection{Ablation Experiment}
\begin{table}[!htpb]
   \centering
   
\begin{tabular}{ccccccc}
\hline
% Group && Accuracy & Precision&Recall&mA \\
AG  &AML&G2FF (CA)&G2FF (SA)&G2FF (CA-SA)& Accuracy &mA \\
\hline
\XSolidBrush  &\XSolidBrush&\XSolidBrush&\XSolidBrush&\XSolidBrush&90.91&88.01 \\
\Checkmark  &\XSolidBrush&\XSolidBrush&\XSolidBrush&\XSolidBrush&91.10&88.16  \\

\Checkmark &\Checkmark&\XSolidBrush&\XSolidBrush&\XSolidBrush&92.22&89.24  \\
\hline
\Checkmark &\Checkmark&\Checkmark&\XSolidBrush&\XSolidBrush&\textbf{92.30}& \textbf{89.33}  \\
\Checkmark &\Checkmark&\XSolidBrush&\Checkmark&\XSolidBrush&92.25& 89.27  \\
\Checkmark &\Checkmark&\XSolidBrush&\XSolidBrush&\Checkmark&92.30& 89.29  \\
\hline
\end{tabular}
\caption{Ablation studies on MGMTN, highlighting the effects of including or excluding AG (Attribute Grouping), AML (Adaptive Mask Learning), and G2FF (Group-Global Feature Fusion). For G2FF, CA denotes Channel Attention and SA denotes Spatial Attention. Experiments were performed on the CelebA dataset.}

\label{tab:4}%
\end{table}

To further investigate the contributions of AML and G2FF within MGMTN, we compare the recognition performance of group features enhanced by AML against that of the global features. In addition, to assess the impact of different attention mechanisms on the proposed framework, we conduct experiments incorporating Channel Attention~\cite{hu2018squeeze}, Spatial Attention~\cite{yan2019stat}, and Cascaded Channel–Spatial Attention~\cite{liu2021global} modules.

The results presented in Table \ref{tab:4} indicate a marked enhancement in the performance of MGMTN relative to the baseline model. Specifically, the accuracy and mA measures show respective improvements of 1.53\% and 1.50\% compared to the baseline. 
The AML method, utilizing only feature ${\rm{\tilde{Gr}}}_4$ improves the baseline by 1.15\% and 1.35\% in accuracy and mA, respectively. Further performance enhancement is observed when both AML and G2FF are employed. 
This demonstrates that fusing local group features extracted via the AML with global features can further enhance the performance of the proposed method. The results indicate that the mA metric of G2FF (CA-SA) exhibits slightly lower performance compared to G2FF (SA). We observe that the enhancement from attention mechanisms on overall performance is marginal. We hypothesize that multi-task networks can effectively learn group features, and overly complex attention mechanisms may lead to parameter redundancy, subsequently causing negative transfer.

{\bf Effectiveness of Adaptive Masks Learning.}
We have analyzed the attribute recognition metrics for feature ${\rm{\tilde{Gr}}}_4$ obtained using AML, and the results are shown in Table \ref{tab:5-1}.

AML enables group features to focus on specific regions within the input image, capturing essential local features for precise identification. This focused approach proves more effective than relying solely on global features, which treat the entire feature map uniformly.
% As a result, this improves recognition performance, especially in handling complex scenes or multi-label learning tasks.

\begin{figure}[!htpb]
\centering
\includegraphics[width=0.8\linewidth]{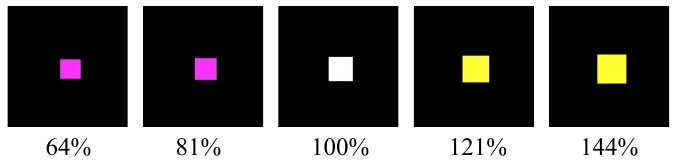}
\caption{\label{fig:8-1}Comparison across different mask region percentages.}
\end{figure}

Adjustments to the size of the predicted mask's region, referencing the scaling shown in Figure \ref{fig:8-1}, were explored for their effect on overall performance, as indicated in Table \ref{tab:5-1}. 
When the mask ratio is increased to 121\% and 144\% compared to the prediction, there is no significant decline in model performance.
Even reducing the mask region to 64\% resulted in a minimal decrease in the accuracy, and mA metrics by 0.2\%, and 0.5\%, respectively, in contrast to the full mask region. 
AML can maintain high predictive performance even in the presence of errors in segmenting group attribute foreground regions, further demonstrating the robustness of the masking method.
% Meanwhile, alternative methods saw a performance decrease of less than 0.2\%. 
\begin{table}[!htpb]
   \centering
     % \resizebox{0.29\textwidth}{!}{
\begin{tabular}{ccc}
\hline
% Group && Accuracy & Precision&Recall&mA \\
 Methods&Accuracy&mA \\
\hline
 64\% of PMR&91.78&88.79\\
  81\% of PMR&91.90&89.15\\
  100\% of PMR&\textbf{91.96}&\textbf{89.20}\\
  121\% of PMR&\textbf{91.96}&89.19\\
  144\% of PMR&91.90&89.10\\

\hline

\end{tabular}
% }
\caption{Comparison of experimental results across different mask ranges relative to the original PMR within the AML framework. Here, PMR refers to the Predicted Mask Region.}

\label{tab:5-1}%
\end{table}
Alterations in the mask region's size did not affect the centrally located feature region, thus preventing a significant drop in model performance and demonstrating the robustness of the AML approach.

\begin{figure}[!htpb]
\centering
\includegraphics[width=0.75\linewidth]{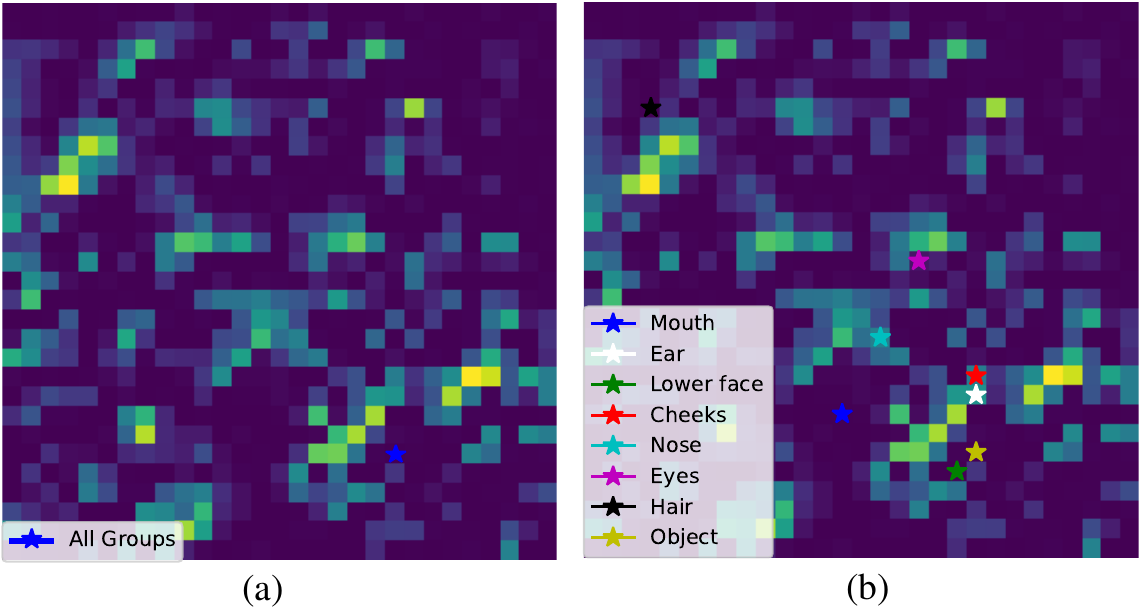}
\caption{\label{fig:10-1}A comparative analysis of AML activation patterns on feature maps. (a) In global-based methods, activations are primarily concentrated within a single feature region, reflecting a one-dimensional response. (b) In contrast, AML produces activation regions that are distinctly aligned with individual group areas, highlighting its capability to model multi-feature and context-aware representations.}
\end{figure}

We visualized the activation spatial positions after max-pooling the group features and global feature maps, as shown in Figure \ref{fig:10-1}. 
The positions obtained after max-pooling the global feature maps are the same for each attribute group and cannot serve as representative features for an attribute group. On the other hand, the spatial positions obtained by max-pooling the group features generated by the AML method are group-specific, indicating that AML can learn effective group features. This further demonstrates the importance of effective feature localization in the process of multi-task prediction.
% 我们针对组特征和全局特征图最大池化后的激活空域位置做了可视化展示，如图2所示。
% 在使用全局特征图最大池化后的位置对于每个属性组都是一样的，并且不能够作为一个属性组的代表性特征。
% 而使用AML方法得到的组特征进行最大池化的空域位置都是与组相关的，这表明AML可以学习到有效组特征。
% 这进一步说明在进行多任务预测的过程中，有效地特征定位是非常有掉的。

% Moreover, the recognition performance achieved by these methods surpassed that of the baseline model, further illustrating that AML has the capability to pinpoint the feature extraction regions corresponding to the attribute groups, thereby augmenting recognition performance. 
% The expansion or contraction of the mask region did not impact the feature region situated in the center, thereby averting a substantial decline in model performance and affirming the resilience of the AML method.
\begin{figure}[!htpb]
\centering
\includegraphics[width=0.55\linewidth]{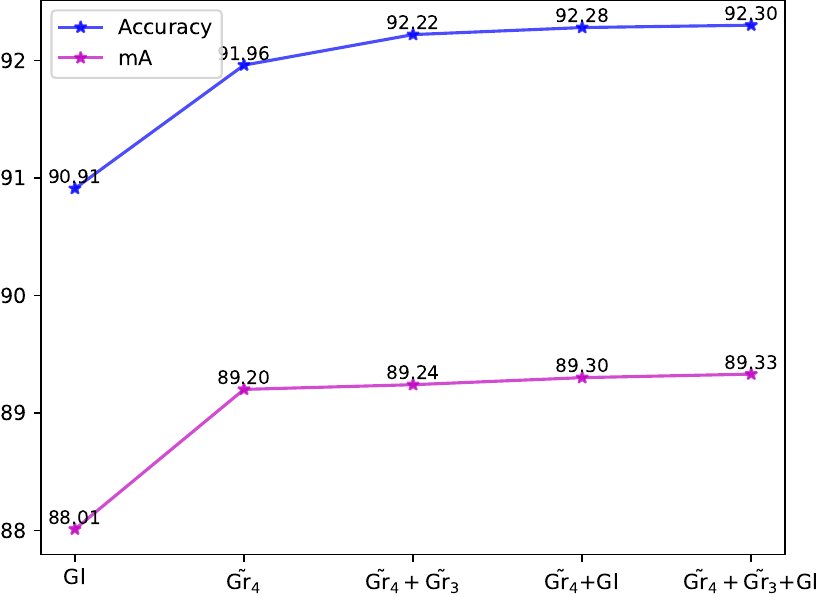}
\caption{\label{fig:9-1}Ablation study of the G2FF method, comparing the experimental results of fusing group features extracted via AML with global features on the CelebA dataset.}
\end{figure}
{\bf Effectiveness of Group-Global Feature Fusion.}
We evaluated the attribute recognition performance obtained through the G2FF module, with the results presented in Figure~\ref{fig:9-1}.
When employing the AML module to extract group features ${\rm{\tilde Gr}}_4$, the model achieved a 1.15\% improvement in accuracy and a 1.35\% gain in mA compared to using the global feature $\rm{Gl}$.
Moreover, combining ${\rm{\tilde Gr}}_4$ with ${\rm{\tilde Gr}}_3$ through AML yielded further performance gains relative to using ${\rm{\tilde Gr}}_4$ alone.
These results demonstrate that the group features extracted by AML effectively capture the attribute characteristics of corresponding regions, thereby enhancing overall model performance.
Finally, the G2FF mechanism fuses group and global features after max-pooling, enabling more effective and comprehensive feature utilization.

\section{Conclusions and Future Work}
We introduced an MGMTN that leverages AML to direct group-specific modules toward accurately predicting regions associated with specific attributes. Our study methodically evaluates the efficacy of each classification head across both global and local feature regions. Moreover, we apply a G2FF technique to normalize the feature distribution, addressing the diverse feature extraction regions related to groups of face attributes. In a visually grounded analysis, our MGMTN approach leverages mask representations to achieve precise localization even under rotated perspectives, thereby validating the effectiveness of utilizing prior group spatial knowledge in FAR. However, in real-world deployment scenarios involving extreme facial poses, occlusions, or noisy conditions, inaccurately localized masks may introduce errors to the classifier and consequently degrade overall performance.

Looking forward, our future work will focus on enhancing performance in real-world deployment scenarios, particularly under extreme facial poses and noisy environments. To this end, we plan to explore innovative techniques such as mask reconstruction and noise injection, investigating their potential to improve attribute classification efficiency through these mechanisms. This line of research aims not only to optimize performance but also to achieve a balanced trade-off between accuracy and robustness.

% \bmhead{Acknowledgements}
% This study was partially supported by National Key Research and Development Program of China (Grant Number 2022YFB330570) and Science and Technology Innovation Plan of Shanghai Science and Technology Commission (Grant Number 21511104302).
\section*{Declarations}

\bmhead{Competing interests}
The authors have no competing interests to declare that are relevant to the content of this article.
% \bmhead{Authors contribution statement}
% Gong Gao is responsible for algorithm implementation, preliminary draft writing and experiment, Xianhui Liu is responsible for supervision and experimental verification, Zhao Jian and Ziqi Xie are responsible for preliminary draft writing, Weidong Zhao is responsible for supervision and Zekai Wang is responsible for paper writing.
\bmhead{Ethical and informed consent for data used}
This article does not contain studies with human participants or animals. As such, informed consent forms are not applicable to this article.
\bmhead{Data availability and access}
The code used or analysed during the current study is available from the corresponding author on reasonable request.
\newpage

\bibliography{sn-bibliography}

@inproceedings{suo2024knowledge,
  title={Knowledge-enhanced dual-stream zero-shot composed image retrieval},
  author={Suo, Yucheng and Ma, Fan and Zhu, Linchao and Yang, Yi},
  booktitle={Proceedings of the IEEE/CVF conference on computer vision and pattern recognition},
  pages={26951--26962},
  year={2024}
}

@article{ahmed2025multi,
  title={Multi-task model with attribute-specific heads for person re-identification},
  author={Ahmed, Md Foysal and Oyshee, Adiba An Nur},
  journal={Pattern Analysis and Applications},
  volume={28},
  number={1},
  pages={38},
  year={2025},
  publisher={Springer}
}

@article{ding2025stable,
  title={Stable attribute group editing for reliable few-shot image generation},
  author={Ding, Guanqi and Han, Xinzhe and Wang, Shuhui and Jin, Xin and Huang, Qingming},
  journal={IEEE Transactions on Circuits and Systems for Video Technology},
  year={2025},
  publisher={IEEE}
}

@inproceedings{wu2025dynamic,
  title={Dynamic modeling of patients, modalities and tasks via multi-modal multi-task mixture of experts},
  author={Wu, Chenwei and Shuai, Zitao and Tang, Zhengxu and Wang, Luning and Shen, Liyue},
  booktitle={The Thirteenth International Conference on Learning Representations},
  year={2025}
}

@inproceedings{hekimoglu2024active,
  title={Active Learning with Task Consistency and Diversity in Multi-Task Networks},
  author={Hekimoglu, Aral and Schmidt, Michael and Marcos-Ramiro, Alvaro},
  booktitle={Proceedings of the IEEE/CVF Winter Conference on Applications of Computer Vision},
  pages={2503--2512},
  year={2024}
}

@inproceedings{lu2017fully,
  title={Fully-adaptive feature sharing in multi-task networks with applications in person attribute classification},
  author={Lu, Yongxi and Kumar, Abhishek and Zhai, Shuangfei and Cheng, Yu and Javidi, Tara and Feris, Rogerio},
  booktitle={Proceedings of the IEEE conference on computer vision and pattern recognition},
  pages={5334--5343},
  year={2017}
}

@inproceedings{chen2021improving,
  title={Improving facial attribute recognition by group and graph learning},
  author={Chen, Zhenghao and Gu, Shuhang and Zhu, Feng and Xu, Jing and Zhao, Rui},
  booktitle={2021 IEEE International Conference on Multimedia and Expo (ICME)},
  pages={1--6},
  year={2021},
  organization={IEEE}
}

@article{chen2023learning,
  title={Learning an attention-aware parallel sharing network for facial attribute recognition},
  author={Chen, Si and Lai, Xinyu and Yan, Yan and Wang, Da-Han and Zhu, Shunzhi},
  journal={Journal of Visual Communication and Image Representation},
  volume={90},
  pages={103745},
  year={2023},
  publisher={Elsevier}
}

@inproceedings{song2022prior,
  title={Prior-Guided Multi-scale Fusion Transformer for Face Attribute Recognition},
  author={Song, Shaoheng and Huang, Huaibo and Wang, Jiaxiang and Zheng, Aihua and He, Ran},
  booktitle={Pattern Recognition and Computer Vision: 5th Chinese Conference, PRCV 2022, Shenzhen, China, November 4--7, 2022, Proceedings, Part I},
  pages={645--659},
  year={2022},
  organization={Springer}
}

@article{li2021auto,
  title={Auto-FERNet: A facial expression recognition network with architecture search},
  author={Li, Shiqian and Li, Wei and Wen, Shiping and Shi, Kaibo and Yang, Yin and Zhou, Pan and Huang, Tingwen},
  journal={IEEE Transactions on Network Science and Engineering},
  volume={8},
  number={3},
  pages={2213--2222},
  year={2021},
  publisher={IEEE}
}

@INPROCEEDINGS{Chennas,
  author={Chen, Mingzhe and Xiao, Xi and Zhang, Bin and Liu, Xinyu and Lu, Runiu},
  booktitle={2022 26th International Conference on Pattern Recognition (ICPR)}, 
  title={Neural Architecture Searching for Facial Attributes-based Depression Recognition}, 
  year={2022},
  volume={},
  number={},
  pages={877-884},
  doi={10.1109/ICPR56361.2022.9956537}}

@inproceedings{fang2017dynamic,
  title={Dynamic Multi-Task Learning with Convolutional Neural Network.},
  author={Fang, Yuchun and Ma, Zhengyan and Zhang, Zhaoxiang and Zhang, Xu-Yao and Bai, Xiang and others},
  booktitle={IJCAI},
  pages={1668--1674},
  year={2017}
}

@inproceedings{zhang2014panda,
  title={Panda: Pose aligned networks for deep attribute modeling},
  author={Zhang, Ning and Paluri, Manohar and Ranzato, Marc'Aurelio and Darrell, Trevor and Bourdev, Lubomir},
  booktitle={Proceedings of the IEEE conference on computer vision and pattern recognition},
  pages={1637--1644},
  year={2014}
}

@inproceedings{hand2017attributes,
  title={Attributes for improved attributes: A multi-task network utilizing implicit and explicit relationships for facial attribute classification},
  author={Hand, Emily and Chellappa, Rama},
  booktitle={Proceedings of the AAAI conference on artificial intelligence},
  volume={31},
  number={1},
  year={2017}
}

@inproceedings{hu2024perspective,
  title={Perspective+ unet: Enhancing segmentation with bi-path fusion and efficient non-local attention for superior receptive fields},
  author={Hu, Jintong and Chen, Siyan and Pan, Zhiyi and Zeng, Sen and Yang, Wenming},
  booktitle={International Conference on Medical Image Computing and Computer-Assisted Intervention},
  pages={499--509},
  year={2024},
  organization={Springer}
}

@article{liu2024differentiable,
  title={Differentiable model scaling using differentiable topk},
  author={Liu, Kai and Wang, Ruohui and Gao, Jianfei and Chen, Kai},
  journal={arXiv preprint arXiv:2405.07194},
  year={2024}
}

@article{wei2025modeling,
  title={Modeling multi-task model merging as adaptive projective gradient descent},
  author={Wei, Yongxian and Tang, Anke and Shen, Li and Hu, Zixuan and Yuan, Chun and Cao, Xiaochun},
  journal={arXiv preprint arXiv:2501.01230},
  year={2025}
}

@inproceedings{rudd2016moon,
  title={Moon: A mixed objective optimization network for the recognition of facial attributes},
  author={Rudd, Ethan M and G{\"u}nther, Manuel and Boult, Terrance E},
  booktitle={Computer Vision--ECCV 2016: 14th European Conference, Amsterdam, The Netherlands, October 11-14, 2016, Proceedings, Part V 14},
  pages={19--35},
  year={2016},
  organization={Springer}
}

@article{alturki2022depth,
  title={Depth-adaptive deep neural network based on learning layer relevance weights},
  author={Alturki, Arwa and Bchir, Ouiem and Ben Ismail, Mohamed Maher},
  journal={Applied Sciences},
  volume={13},
  number={1},
  pages={398},
  year={2022},
  publisher={MDPI}
}

@article{wang2024advances,
  title={Advances in neural architecture search},
  author={Wang, Xin and Zhu, Wenwu},
  journal={National Science Review},
  volume={11},
  number={8},
  pages={nwae282},
  year={2024},
  publisher={Oxford University Press}
}

@article{han2017heterogeneous,
  title={Heterogeneous face attribute estimation: A deep multi-task learning approach},
  author={Han, Hu and Jain, Anil K and Wang, Fang and Shan, Shiguang and Chen, Xilin},
  journal={IEEE transactions on pattern analysis and machine intelligence},
  volume={40},
  number={11},
  pages={2597--2609},
  year={2017},
  publisher={IEEE}
}

@inproceedings{cao2018partially,
  title={Partially shared multi-task convolutional neural network with local constraint for face attribute learning},
  author={Cao, Jiajiong and Li, Yingming and Zhang, Zhongfei},
  booktitle={Proceedings of the IEEE Conference on computer vision and pattern recognition},
  pages={4290--4299},
  year={2018}
}

@inproceedings{shu2021learning,
  title={Learning spatial-semantic relationship for facial attribute recognition with limited labeled data},
  author={Shu, Ying and Yan, Yan and Chen, Si and Xue, Jing-Hao and Shen, Chunhua and Wang, Hanzi},
  booktitle={Proceedings of the IEEE/CVF Conference on Computer Vision and Pattern Recognition},
  pages={11916--11925},
  year={2021}
}

@inproceedings{zhang2022resnest,
  title={Resnest: Split-attention networks},
  author={Zhang, Hang and Wu, Chongruo and Zhang, Zhongyue and Zhu, Yi and Lin, Haibin and Zhang, Zhi and Sun, Yue and He, Tong and Mueller, Jonas and Manmatha, R and others},
  booktitle={Proceedings of the IEEE/CVF conference on computer vision and pattern recognition},
  pages={2736--2746},
  year={2022}
}

@article{mao2020deep,
  title={Deep multi-task multi-label CNN for effective facial attribute classification},
  author={Mao, Longbiao and Yan, Yan and Xue, Jing-Hao and Wang, Hanzi},
  journal={IEEE Transactions on Affective Computing},
  volume={13},
  number={2},
  pages={818--828},
  year={2020},
  publisher={IEEE}
}

@inproceedings{liu2015deep,
  title={Deep learning face attributes in the wild},
  author={Liu, Ziwei and Luo, Ping and Wang, Xiaogang and Tang, Xiaoou},
  booktitle={Proceedings of the IEEE international conference on computer vision},
  pages={3730--3738},
  year={2015}
}

@inproceedings{huang2008labeled,
  title={Labeled faces in the wild: A database forstudying face recognition in unconstrained environments},
  author={Huang, Gary B and Mattar, Marwan and Berg, Tamara and Learned-Miller, Eric},
  booktitle={Workshop on faces in'Real-Life'Images: detection, alignment, and recognition},
  year={2008}
}

@inproceedings{fang2023eva,
  title={Eva: Exploring the limits of masked visual representation learning at scale},
  author={Fang, Yuxin and Wang, Wen and Xie, Binhui and Sun, Quan and Wu, Ledell and Wang, Xinggang and Huang, Tiejun and Wang, Xinlong and Cao, Yue},
  booktitle={Proceedings of the IEEE/CVF Conference on Computer Vision and Pattern Recognition},
  pages={19358--19369},
  year={2023}
}

@article{wang2023masked,
  title={Masked face recognition dataset and application},
  author={Wang, Zhongyuan and Huang, Baojin and Wang, Guangcheng and Yi, Peng and Jiang, Kui},
  journal={IEEE Transactions on Biometrics, Behavior, and Identity Science},
  year={2023},
  publisher={IEEE}
}

@inproceedings{hu2018squeeze,
  title={Squeeze-and-excitation networks},
  author={Hu, Jie and Shen, Li and Sun, Gang},
  booktitle={Proceedings of the IEEE conference on computer vision and pattern recognition},
  pages={7132--7141},
  year={2018}
}

@article{paszke2019pytorch,
  title={Pytorch: An imperative style, high-performance deep learning library},
  author={Paszke, Adam and Gross, Sam and Massa, Francisco and Lerer, Adam and Bradbury, James and Chanan, Gregory and Killeen, Trevor and Lin, Zeming and Gimelshein, Natalia and Antiga, Luca and others},
  journal={Advances in neural information processing systems},
  volume={32},
  year={2019}
}

@ARTICLE{10064142,
  author={Chen, Si and Zhu, Xueyan and Yan, Yan and Zhu, Shunzhi and Li, Shao-Zi and Wang, Da-Han},
  journal={IEEE Transactions on Circuits and Systems for Video Technology},
  title={Identity-Aware Contrastive Knowledge Distillation for Facial Attribute Recognition},
  year={2023},
  volume={},
  number={},
  pages={1-1},
  doi={10.1109/TCSVT.2023.3253799}}

@article{zheng2020blan,
  title={BLAN: Bi-directional ladder attentive network for facial attribute prediction},
  author={Zheng, Xin and Huang, Huaibo and Guo, Yanqing and Wang, Bo and He, Ran},
  journal={Pattern Recognition},
  volume={100},
  pages={107155},
  year={2020},
  publisher={Elsevier}
}

@inproceedings{wang2019dynamic,
  title={Dynamic curriculum learning for imbalanced data classification},
  author={Wang, Yiru and Gan, Weihao and Yang, Jie and Wu, Wei and Yan, Junjie},
  booktitle={Proceedings of the IEEE/CVF international conference on computer vision},
  pages={5017--5026},
  year={2019}
}

@inproceedings{zheng2022general,
  title={General facial representation learning in a visual-linguistic manner},
  author={Zheng, Yinglin and Yang, Hao and Zhang, Ting and Bao, Jianmin and Chen, Dongdong and Huang, Yangyu and Yuan, Lu and Chen, Dong and Zeng, Ming and Wen, Fang},
  booktitle={Proceedings of the IEEE/CVF Conference on Computer Vision and Pattern Recognition},
  pages={18697--18709},
  year={2022}
}

@article{kingma2014adam,
  title={Adam: A method for stochastic optimization},
  author={Kingma, Diederik P and Ba, Jimmy},
  journal={arXiv preprint arXiv:1412.6980},
  year={2014}
}

@article{jia2023tfgnet,
  title={TFGNet: Traffic Salient Object Detection Using a Feature Deep Interaction and Guidance Fusion},
  author={Jia, Ning and Sun, Yougang and Liu, Xianhui},
  journal={IEEE Transactions on Intelligent Transportation Systems},
  year={2023},
  publisher={IEEE}
}

@inproceedings{zhou2023feature,
  title={Feature decomposition for reducing negative transfer: a novel multi-task learning method for recommender system (student abstract)},
  author={Zhou, Jie and Yu, Qian and Luo, Chuan and Zhang, Jing},
  booktitle={Proceedings of the AAAI Conference on Artificial Intelligence},
  volume={37},
  number={13},
  pages={16390--16391},
  year={2023}
}

@inproceedings{li2022equalized,
  title={Equalized focal loss for dense long-tailed object detection},
  author={Li, Bo and Yao, Yongqiang and Tan, Jingru and Zhang, Gang and Yu, Fengwei and Lu, Jianwei and Luo, Ye},
  booktitle={Proceedings of the IEEE/CVF Conference on Computer Vision and Pattern Recognition},
  pages={6990--6999},
  year={2022}
}

@inproceedings{he2018harnessing,
  title={Harnessing Synthesized Abstraction Images to Improve Facial Attribute Recognition.},
  author={He, Keke and Fu, Yanwei and Zhang, Wuhao and Wang, Chengjie and Jiang, Yu-Gang and Huang, Feiyue and Xue, Xiangyang},
  booktitle={IJCAI},
  pages={733--740},
  year={2018}
}

@inproceedings{he2019mtcnn,
  title={Mtcnn with weighted loss penalty and adaptive threshold learning for facial attribute prediction},
  author={He, Xingting and Wang, Pingyu and Zhao, Zhicheng and Zhao, Yanyun and Su, Fei},
  booktitle={2019 IEEE International Conference on Multimedia \& Expo Workshops (ICMEW)},
  pages={180--185},
  year={2019},
  organization={IEEE}
}

@inproceedings{lingenfelter2021improving,
  title={Improving evaluation of facial attribute prediction models},
  author={Lingenfelter, Bryson and Hand, Emily M},
  booktitle={2021 16th IEEE International Conference on Automatic Face and Gesture Recognition (FG 2021)},
  pages={1--7},
  year={2021},
  organization={IEEE}
}

@article{li2024cspformer,
  title={CSPFormer: A cross-spatial pyramid transformer for visual place recognition},
  author={Li, Zhenyu and Xu, Pengjie},
  journal={Neurocomputing},
  pages={127472},
  year={2024},
  publisher={Elsevier}
}

@article{wu2024attributes,
  title={Attributes-Assisted Joint Contrastive Learning for Person Re-Identification},
  author={Wu, Qingru and Zhou, Zhiheng and Niu, Chang and Liu, Xiaosheng and Tao, Xiyuan and Li, Bo},
  journal={IEEE Internet of Things Journal},
  year={2024},
  publisher={IEEE}
}

@inproceedings{kalayeh2017improving,
  title={Improving facial attribute prediction using semantic segmentation},
  author={Kalayeh, Mahdi M and Gong, Boqing and Shah, Mubarak},
  booktitle={Proceedings of the IEEE conference on computer vision and pattern recognition},
  pages={6942--6950},
  year={2017}
}

@inproceedings{ding2018deep,
  title={A deep cascade network for unaligned face attribute classification},
  author={Ding, Hui and Zhou, Hao and Zhou, Shaohua and Chellappa, Rama},
  booktitle={Proceedings of the AAAI Conference on Artificial Intelligence},
  volume={32},
  number={1},
  year={2018}
}

@inproceedings{ehrlich2016facial,
  title={Facial attributes classification using multi-task representation learning},
  author={Ehrlich, Max and Shields, Timothy J and Almaev, Timur and Amer, Mohamed R},
  booktitle={Proceedings of the IEEE Conference on Computer Vision and Pattern Recognition Workshops},
  pages={47--55},
  year={2016}
}

@article{cargan2024local,
  title={Local-global methods for generalised solar irradiance forecasting},
  author={Cargan, Timothy R and Landa-Silva, Dario and Triguero, Isaac},
  journal={Applied Intelligence},
  volume={54},
  number={2},
  pages={2225--2247},
  year={2024},
  publisher={Springer}
}

@inproceedings{kobayashi2019global,
  title={Global feature guided local pooling},
  author={Kobayashi, Takumi},
  booktitle={Proceedings of the IEEE/CVF International Conference on Computer Vision},
  pages={3365--3374},
  year={2019}
}

@article{xie2023farp,
  title={FARP-Net: Local-global feature aggregation and relation-aware proposals for 3D object detection},
  author={Xie, Tao and Wang, Li and Wang, Ke and Li, Ruifeng and Zhang, Xinyu and Zhang, Haoming and Yang, Linqi and Liu, Huaping and Li, Jun},
  journal={IEEE Transactions on Multimedia},
  year={2023},
  publisher={IEEE}
}

@article{yan2019stat,
  title={STAT: Spatial-temporal attention mechanism for video captioning},
  author={Yan, Chenggang and Tu, Yunbin and Wang, Xingzheng and Zhang, Yongbing and Hao, Xinhong and Zhang, Yongdong and Dai, Qionghai},
  journal={IEEE transactions on multimedia},
  volume={22},
  number={1},
  pages={229--241},
  year={2019},
  publisher={IEEE}
}

@article{liu2021global,
  title={Global attention mechanism: Retain information to enhance channel-spatial interactions},
  author={Liu, Yichao and Shao, Zongru and Hoffmann, Nico},
  journal={arXiv preprint arXiv:2112.05561},
  year={2021}
}

@inproceedings{dalal2005histograms,
  title={Histograms of oriented gradients for human detection},
  author={Dalal, Navneet and Triggs, Bill},
  booktitle={2005 IEEE computer society conference on computer vision and pattern recognition (CVPR'05)},
  volume={1},
  pages={886--893},
  year={2005},
  organization={Ieee}
}

@article{lowe2004distinctive,
  title={Distinctive image features from scale-invariant keypoints},
  author={Lowe, David G},
  journal={International journal of computer vision},
  volume={60},
  number={2},
  pages={91--110},
  year={2004},
  publisher={Springer}
}

@inproceedings{ojala1994performance,
  title={Performance evaluation of texture measures with classification based on Kullback discrimination of distributions},
  author={Ojala, Timo and Pietikainen, Matti and Harwood, David},
  booktitle={Proceedings of 12th international conference on pattern recognition},
  volume={1},
  pages={582--585},
  year={1994},
  organization={IEEE}
}

@article{cortes1995support,
  title={Support-vector networks},
  author={Cortes, Corinna and Vapnik, Vladimir},
  journal={Machine learning},
  volume={20},
  number={3},
  pages={273--297},
  year={1995},
  publisher={Springer}
}

@article{ghani2024securing,
  title={Securing synthetic faces: A GAN-blockchain approach to privacy-enhanced facial recognition},
  author={Ghani, Muhammad Ahmad Nawaz Ul and She, Kun and Rauf, Muhammad Arslan and Alajmi, Masoud and Ghadi, Yazeed Yasin and Algarni, Abdulmohsen},
  journal={Journal of King Saud University-Computer and Information Sciences},
  volume={36},
  number={4},
  pages={102036},
  year={2024},
  publisher={Elsevier}
}

@inproceedings{han2024face,
  title={Face-adapter for pre-trained diffusion models with fine-grained id and attribute control},
  author={Han, Yue and Zhu, Junwei and He, Keke and Chen, Xu and Ge, Yanhao and Li, Wei and Li, Xiangtai and Zhang, Jiangning and Wang, Chengjie and Liu, Yong},
  booktitle={European Conference on Computer Vision},
  pages={20--36},
  year={2024},
  organization={Springer}
}

@inproceedings{huang2024attribute,
  title={Attribute-guided pedestrian retrieval: Bridging person re-id with internal attribute variability},
  author={Huang, Yan and Zhang, Zhang and Wu, Qiang and Zhong, Yi and Wang, Liang},
  booktitle={Proceedings of the IEEE/CVF conference on computer vision and pattern recognition},
  pages={17689--17699},
  year={2024}
}

@article{gao2025far,
  title={FAR-AMTN: Attention Multi-Task Network for Face Attribute Recognition},
  author={Gao, Gong and Wang, Zekai and Liu, Xianhui and Zhao, Weidong},
  journal={Computer Vision and Image Understanding},
  pages={104426},
  year={2025},
  publisher={Elsevier}
}

\end{document}